\documentclass[sigconf]{acmart}
\acmConference[ ]{ }{ }{ }
\begin{document}

\title{DiffusionCinema: Text-to-Aerial Cinematography}


\author{Valerii Serpiva}
\authornote{Both authors contributed equally to this research.}
\email{Valerii.Serpiva@skoltech.ru}
\orcid{0009-0004-3285-3937}
\affiliation{%
  \institution{Skolkovo Institute of Science and Technology}
  \city{Moscow}
  \country{Russia}
}

\author{Artem Lykov}
\email{Artem.Lykov@skoltech.ru}
\authornotemark[1]
\orcid{0000-0001-6119-2366}
\affiliation{%
  \institution{Skolkovo Institute of Science and Technology}
  \city{Moscow}
  \country{Russia}
}

\author{Jeffrin Sam}
\email{Jeffrin.Sam@skoltech.ru}
\orcid{0009-0000-8635-5379}
\affiliation{%
  \institution{Skolkovo Institute of Science and Technology}
  \city{Moscow}
  \country{Russia}
}

\author{Aleksey Fedoseev}
\email{Aleksey.Fedoseev@skoltech.ru}
\orcid{0000-0003-2506-9111}
\affiliation{%
  \institution{Skolkovo Institute of Science and Technology}
  \city{Moscow}
  \country{Russia}
}

\author{Dzmitry Tsetserukou}
\email{D.Tsetserukou@skoltech.ru}
\orcid{0000-0001-8055-5345}
\affiliation{%
  \institution{Skolkovo Institute of Science and Technology}
  \city{Moscow}
  \country{Russia}
}


\begin{abstract}
We propose a novel Unmanned Aerial Vehicles (UAV) assisted creative capture system that leverages diffusion models to interpret high-level natural language prompts and automatically generate optimal flight trajectories for cinematic video recording. Instead of manually piloting the drone, the user simply describes the desired shot (e.g., “orbit around me slowly from the right and reveal the background waterfall”). Our system encodes the prompt along with an initial visual snapshot from the onboard camera, and a diffusion model samples plausible spatio-temporal motion plans that satisfy both the scene geometry and shot semantics. The generated flight trajectory is then executed autonomously by the UAV to record smooth, repeatable video clips that match the prompt. User evaluation using NASA-TLX showed a significantly lower overall workload with our interface ($M = 21.6$) compared to a traditional remote controller ($M = 58.1$), demonstrating a substantial reduction in perceived effort. Mental demand ($M = 11.5$ vs. $60.5$) and frustration ($M = 14.0$ vs. $54.5$) were also markedly lower for our system, confirming clear usability advantages in autonomous text-driven flight control. This project demonstrates a new interaction paradigm: text-to-cinema flight, where diffusion models act as the “creative operator” converting story intentions directly into aerial motion.
\end{abstract}

\begin{teaserfigure}
\centering
  \includegraphics[width=0.9\textwidth]{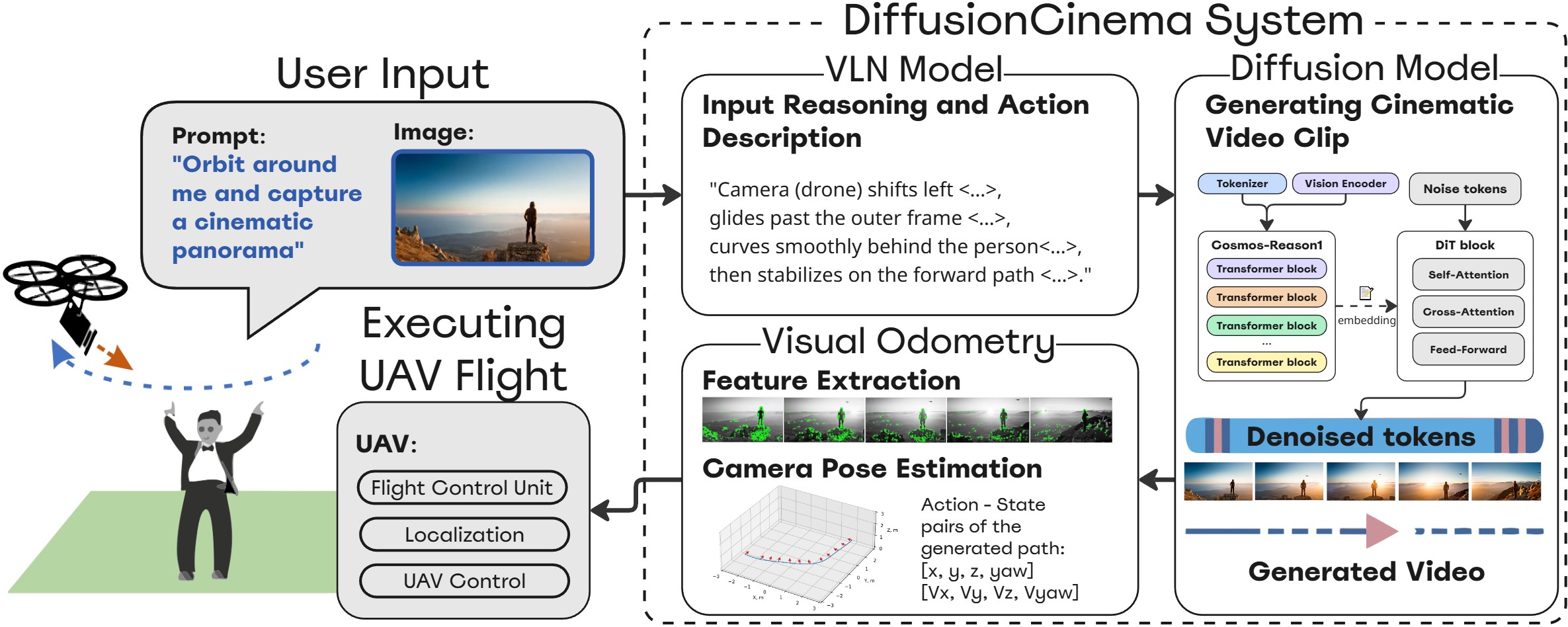}
  \caption{DiffusionCinema converts a user’s text prompt and an initial image into a UAV trajectory for autonomously recording a cinematic video. The VLN model interprets the prompt into flight actions, the diffusion model generates the video sequence, visual odometry extracts the synthetic camera movement, and the UAV follows the planned path.}
  \label{fig:teaser}
\end{teaserfigure}

\received{20 February 2007}
\received[revised]{12 March 2009}
\received[accepted]{5 June 2009}

\maketitle

\section{Introduction}

Autonomous camera drones capable of following high-level verbal instructions and capturing cinematic aerial footage have long been envisioned in science fiction. Today, UAVs are widely used in film and media production \cite{10.1145/3347713}, enabling dynamic viewpoints difficult to achieve from the ground. However, producing high-quality aerial footage remains challenging due to limited time and personnel, while requiring efficient trajectories and consistent, meaningful visual framing.

Currently, this burden falls on skilled First-Person View (FPV) pilots, who navigate drones while framing shots. Mastery of FPV control \cite{9372809} requires extensive training to achieve stable high-speed flight, obstacle avoidance, and accurate 3D positioning, while cinematic storytelling demands expertise in composition, motion, timing, and transitions. Recent advances in AI-driven drone control \cite{10530312, 9729807, 11007799} can automate high-speed maneuvers and complex navigation with minimal pilot input. The combination of these skill sets is rare but essential for professional UAV cinematography \cite{8486586}, making high-end aerial filming largely inaccessible to non-experts despite affordable hardware.

Recent research in language-conditioned UAV control has demonstrated promising mechanisms for abstracting low-level piloting. Prior work includes language-driven planning and control systems \cite{10.1007/978-981-99-6495-6_31,electronics14214312,10.1145/3774655,uav_vla_2501}, Vision Language Action (VLA) models that map visual observations and text directly to control commands \cite{racevla,cognitivedrone,RT1,RT2,OpenVLA}, and aerial Vision Language Navigation (VLN) frameworks that ground instructions in 3D environments \cite{AerialVLN,2410.07087,CityNav,EmbodiedCity,STMR_aerial}. The diffusion model jointly predicts visual traversability and feasible trajectories from a single RGB image, enabling prompt-free, embodiment-agnostic navigation \cite{zhura2025swarmdiffusionendtoendtraversabilityguideddiffusion}. These approaches focus on navigation accuracy, efficiency, or task completion, but not on the stylistic and temporal qualities of cinematic footage. Video diffusion models, trained on large-scale human-shot videos, provide strong generative priors for realistic motion, transitions, and cinematic style. Recent work demonstrates their utility for UAV and robotics control by generating trajectories aligned with realistic drone motion \cite{racevla,cognitivedrone}. 
This makes diffusion models appealing for aerial cinematography, as they encode human-like camera behavior without requiring explicit supervision on filming rules.

By treating the diffusion model as a cinematographic prior within the control loop, DiffusionCinema enables text-driven aerial filming without manual piloting. In contrast to prior language-driven UAV systems focused on navigation or efficiency, our approach explicitly targets cinematic quality by inheriting stylistic camera motion from large-scale human-shot video data. This paradigm substantially lowers the barrier for aerial content creation while providing professionals with a tool for rapid prototyping of complex shots.

\section{Related Work}

\subsection{Language-Driven Aerial Autonomy}

Natural language has emerged as an effective high-level interface for UAV control, abstracting low-level piloting into semantic commands. Early work integrated Large Language Models (LLM) into task planning pipelines that translate verbal goals into executable robotic actions \cite{10.1007/978-981-99-6495-6_31}. This direction evolved into real-time multilingual voice control systems \cite{electronics14214312} and LLM-based reasoning frameworks for adaptive UAV and swarm trajectory planning in dynamic environments \cite{10.1145/3774655}.

Transformer-based vision-language systems further advanced this paradigm. UAV-VLA \cite{uav_vla_2501}, for example, generates large-scale aerial mission plans directly from satellite imagery and textual descriptions, significantly accelerating mission design. While these approaches demonstrate the feasibility and accessibility of language-conditioned UAV control, they primarily produce waypoint-based plans and do not model continuous camera motion or visual style.

\subsection{Vision-Language-Action Models}

VLA models unify perception, language understanding, and control in end-to-end architectures \cite{RT1,RT2,OpenVLA}. In aerial robotics, RaceVLA \cite{racevla} maps FPV observations and language instructions directly to continuous control commands, while CognitiveDrone \cite{cognitivedrone} augments this paradigm with reasoning modules for cognitively complex tasks.

Despite their effectiveness for navigation and task execution, existing VLA models are trained on utilitarian objectives and do not encode cinematographic principles. Consequently, they lack the implicit priors required for expressive camera motion and visual storytelling.

\subsection{Aerial Vision-Language Navigation}

VLN addresses the grounding of natural language in embodied visual environments. AerialVLN \cite{AerialVLN} introduced a benchmark for instruction-following UAVs in large-scale 3D spaces. Subsequent work expanded this setting to city-scale environments \cite{CityNav,EmbodiedCity} and realistic UAV dynamics with multimodal LLM guidance \cite{2410.07087}. Semantic Topo-Metric Representations further enhance spatial reasoning for aerial navigation \cite{STMR_aerial}.

These VLN frameworks emphasize the goal completion and path efficiency, yet they do not model cinematic objectives such as shot continuity, framing, or aesthetic camera motion.

\subsection{Video Diffusion Models as Motion Priors}

Video diffusion models now dominate high-fidelity video generation \cite{videodiffusion_survey}, producing long, coherent sequences with realistic camera motion. Trained on large corpora of human-recorded footage, they implicitly encode cinematographic patterns such as tracking shots, orbits, and reveals.

FlightDiffusion \cite{flightdiffusion} demonstrated that diffusion models can generate FPV video trajectories and associated UAV states as motion priors for control policy learning. DiffusionCinema applies diffusion-based video synthesis as a cinematographic prior for real-time UAV trajectory generation, enabling language-driven aerial cinematography grounded in human-shot videos.

\begin{figure*}[t]
\centering
  \includegraphics[width=1.0\linewidth]{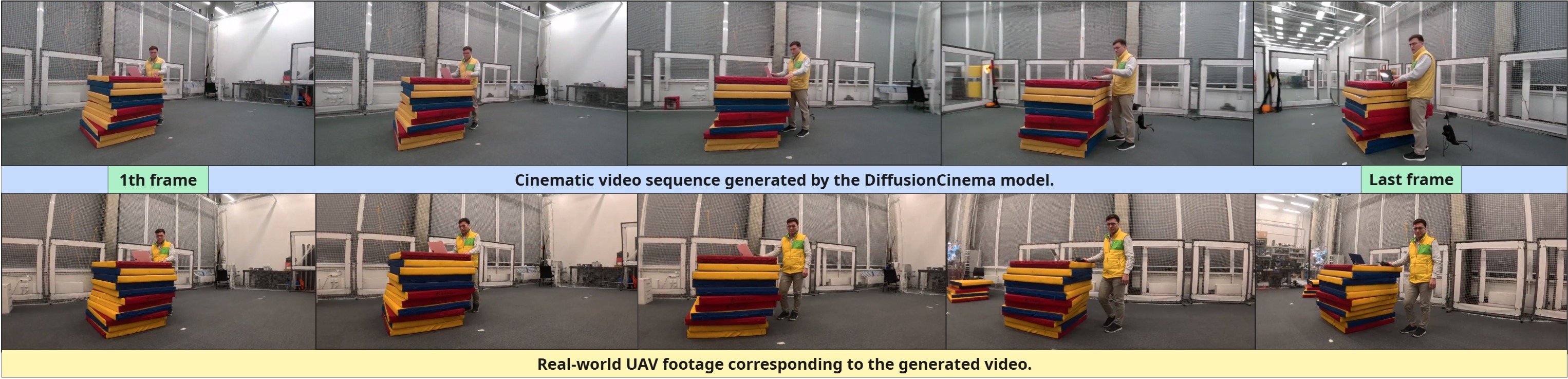}
  \caption{Generated video sequence for the task “Make a cinematic shot for my video vlog with me as the host.” The top row of images shows frames generated by the DiffusionCinema model, while the bottom row shows the corresponding real footage captured by the UAV following the host.}
  \Description{Generated video sequence for the task “Make a cinematic shot for my video vlog with me as the host.” The top row shows frames generated by the DiffusionCinema model, while the bottom row shows the corresponding real footage captured by the UAV following the host.}
  \label{fig:cosmos}
\end{figure*}

\section{System Overview}

\subsection{Diffusion Model}

The depicted architecture illustrates a diffusion-based generation pipeline that fuses multimodal embeddings with transformer-driven denoising to produce high-quality visual outputs conditioned on text and imagery. The system accepts two parallel inputs: a text prompt, tokenized into a linguistic embedding, and an image or video input, encoded into a visual embedding. These are combined in the Cosmos-Reason1 module~\cite{NVIDIA_2025_Cosmos}, a transformer-based multimodal reasoning layer that aligns textual semantics with visual context to form a unified latent representation. This representation is then projected into a modality-compatible embedding that conditions the diffusion stage.

The projected embedding guides a sequence of Diffusion Transformer (DiT) blocks operating over time-dependent latent tokens that represent the target video in a compressed space. Each block includes LayerNorm-modulated Self-Attention, Cross-Attention, and Feed-Forward components. Self-Attention refines internal token structure, while Cross-Attention injects semantic and visual conditioning signals. Time embeddings are added at each step to preserve temporal coherence. Iterative denoising progressively transforms noisy latent tokens into clean representations, effectively decoding the autoregressively predicted sequence. The resulting latent tokens can then be rendered into a final frame sequence or visual output that reflects both the prompt and conditioning imagery.

The hardware configuration consisted of an NVIDIA GeForce RTX 5090 GPU (32GB VRAM) and an Intel Core i9-14900K CPU.  This setup allowed the system to perform multimodal reasoning and video synthesis for a 12 s flight sequence up to 2 min with 480p video resolution.

\subsection{Path Extraction and UAV Flight Execution}

The process begins with the user providing a natural-language text prompt $P$ and an image $I_0$, which together specify both the desired drone behavior and the intended visual style.

A VLN model interprets this input and generates a high-level action description $\mathcal{A} = \{a_1, a_2, \ldots, a_T\}$, where each $a_t$ denotes a semantic flight instruction at time step $t$, defining the UAV's trajectory $\tau = \{x_t, y_t, z_t, \psi_t\}_{t=1}^{T}$. The diffusion model then tokenizes and encodes the visual input into latent tokens $z_0$, and iteratively denoises them according to
\begin{equation}
z_{t-1} = f_\theta(z_t, \epsilon_t, \mathcal{C}), \quad t = T, \ldots, 1,
\end{equation}
where $f_\theta(\cdot)$ is the transformer-based denoising function, $\epsilon_t$ represents sampled noise, and $\mathcal{C}$ is the conditioning derived from $(P, I_0, \mathcal{A})$. The final denoised tokens $z_0$ are decoded into a generated video $V_{\mathrm{syn}}$ that adheres to the cinematic motion defined by the prompt. Leveraging ORB-SLAM3 ~\cite{ORBSLAM3_TRO}, we extract state-action pairs from the synthetic video:
\begin{equation}
\mathcal{S} = \{(s_t, a_t)\}_{t=1}^{T}, \quad s_t = [x_t, y_t, z_t, \psi_t, v_t],
\end{equation}

where $s_t$ and $v_t$ 
denote the estimated UAV state and its velocity, respectively. These pairs enable alignment between virtual scene dynamics and real-world UAV behavior.

During flight, onboard modules manage control, localization, and state estimation, while visual odometry extracts image features $F_t$ and estimates UAV pose via:
\begin{equation}
\hat{s}_t = g(F_t, \hat{s}_{t-1}),
\end{equation}
where $g(\cdot)$ denotes the pose estimation function. This produces a sequence of estimated states $\hat{\tau} = \{\hat{s}_1, \ldots, \hat{s}_T\}$.

To execute the cinematic motion defined by the generated trajectory $\tau$, the UAV follows the desired path using a PID-based controller. For each state component (position or yaw), the control input $u_t$ is computed as:
\begin{equation}
u_t = K_P e_t + K_I \int_{0}^{t} e_\xi \, d\xi + K_D \frac{d e_t}{dt},
\end{equation}
where $e_t = s_t^{\ast} - \hat{s}_t$ is the tracking error between the desired state $s_t^{\ast} \in \tau$ and the estimated state $\hat{s}_t$ obtained by OpenVINS \cite{9196524}, and $K_P$, $K_I$, and $K_D$ denote the proportional, integral, and derivative gains respectively. The result is a generated video $V_{\mathrm{syn}}$ that combines the captured environment with user-defined creative instructions, linking physical UAV motion to virtual cinematic storytelling.

\begin{figure}[t]
\centering
  \includegraphics[width=0.9\linewidth]{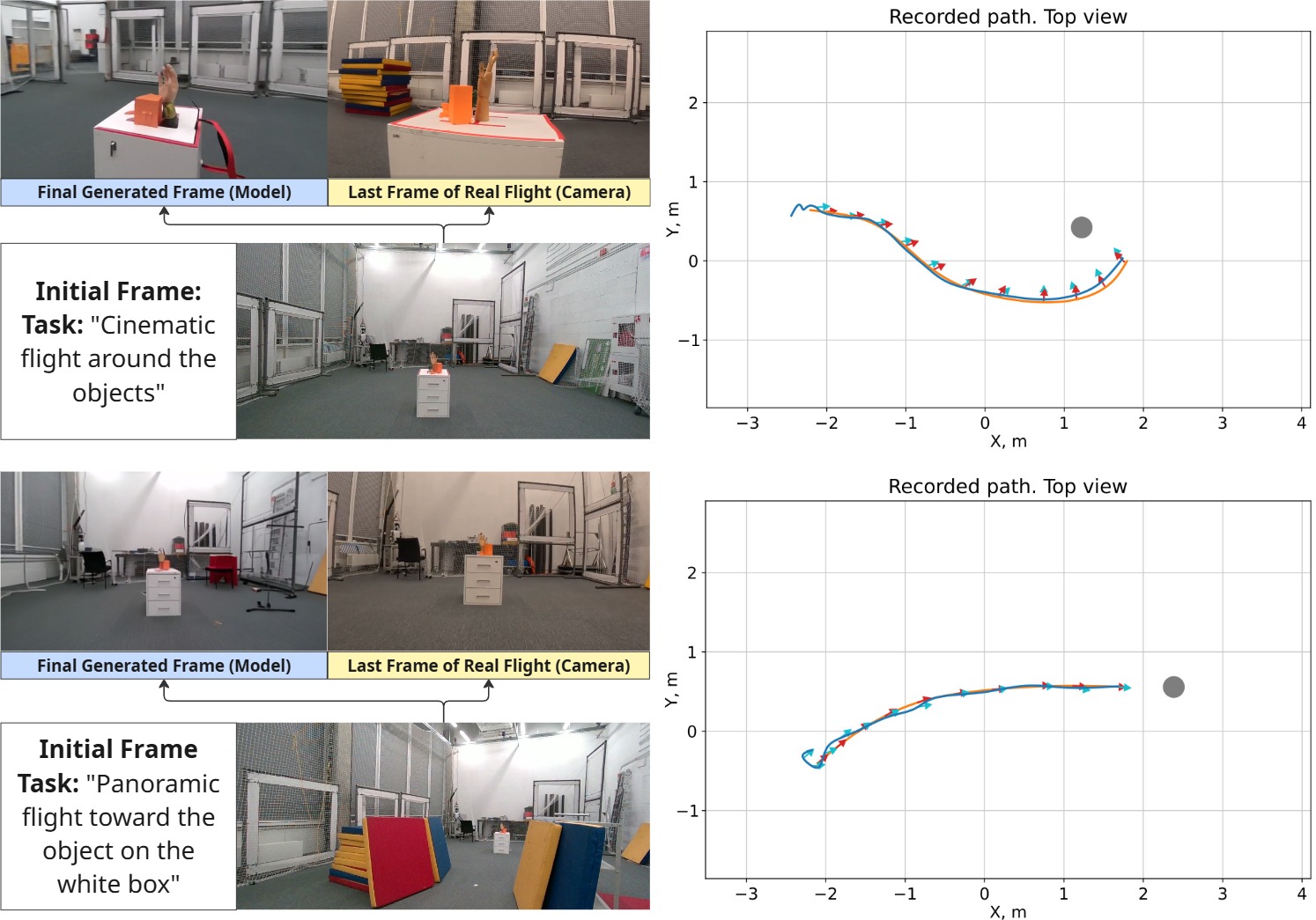}
  \caption{Comparison of the trajectory generated by ORB-SLAM from synthetic video and the trajectory executed by a real UAV: yellow line is the generated trajectory; blue line is the actual UAV path; gray circle-target object.}
  \Description{Experimental comparison of model-generated and real UAV flight paths: the yellow line shows the diffusion model trajectory, the blue line the actual UAV path, and the gray circle the target object position. Left images depict the initial prompt frame, the final generated frame, and the final real flight frame.}
  \label{fig:exp_setup}
\end{figure}

Around $46.7\%$ of generated videos were successful, and ORB-SLAM3 was able to extract trajectories accurately even in the presence of hallucinated objects \cite{flightdiffusion}. The end-to-end pipeline-from prompt to flight start takes approximately 30 s to 2 min. The generated path and the UAV’s executed flight, as well as the recorded and synthetic videos, were compared, as shown in Fig. \ref{fig:exp_setup}.


\section{Preliminary User Evaluation}

\begin{figure}[t]
\centering
  \includegraphics[width=0.9\linewidth]{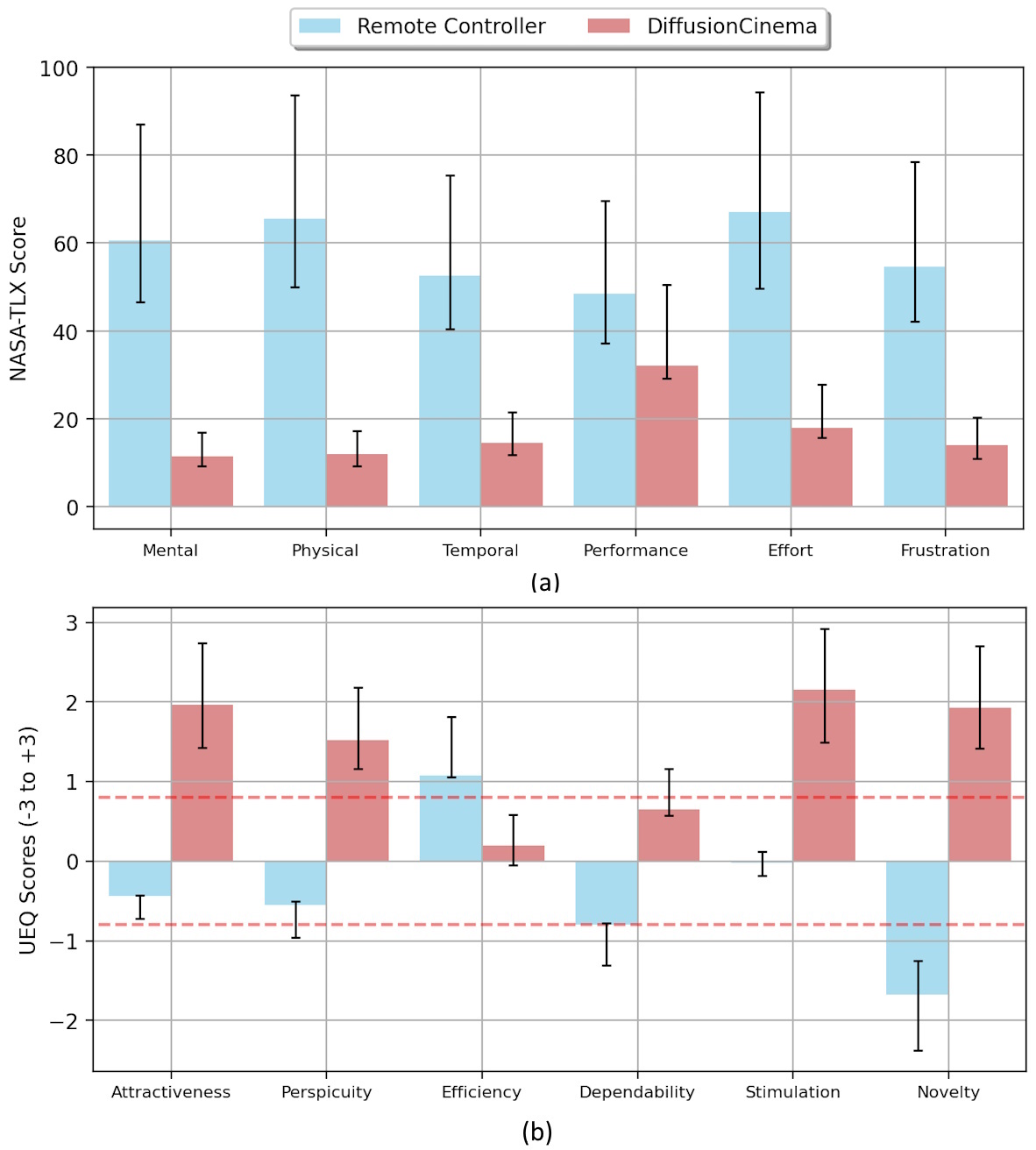}  
  \vspace{-0.5cm}
  \caption{(a) NASA-TLX questionnaire. Black error bars denote 95\% confidence interval (CI). (b) The UEQ questionnaire. Black error bars denote 95\% CI. Values between -0.8 and 0.8 (in the dashed line) represent a neutral evaluation.}
  \Description{eval.}
  \label{fig:NASA}
\end{figure}

\subsection{User Study Setup}

The preliminary study involved a total of 10 participants (6 males and 4 females), aged between 18 and 31 years ($M = 25.9$). Participants showed varied prior experience: with drones, four reported regular use, three occasional use, and three no experience; with VLM systems, five reported regular use, three occasional use, and two no experience. 
Each participant recorded aerial footage with two control interfaces: (1) manual control via a Remote Controller (RC) and (2) automatic execution of a written prompt by DiffusionCinema. The task was to capture three cinematic maneuvers: orbiting, moving toward the object, and pan-orbiting. After each flight, participants completed a user survey consisting of the NASA-TLX \cite{NASA} and the User Experience Questionnaire (UEQ) \cite{UEQ}.

\subsection{User Study Results}

Given the nonparametric nature of the data, a Wilcoxon signed-rank test for paired samples was conducted. As shown in Fig. \ref{fig:NASA}, the NASA-TLX results indicated a statistically significant difference in overall perceived workload between the interfaces ($V = 6.0, p = 0.02$). The RC elicited a higher workload ($M = 58.1, SD = 15.6$) than the proposed interface ($M = 21.6, SD = 10.6$). Analysis of the subscales confirmed our hypothesis that physical demand was rated substantially lower for the DiffusionCinema interface ($M = 12.0, SD = 5.4$), with the difference statistically significant ($V = 6.75, p=8.4\cdot10^{-5}$). The frustration subscale was significantly lower for the DiffusionCinema interface ($M = 14.0, SD = 6.99$) compared to the RC ($M = 54.5, SD = 24.08$) ($V = 1.0, p = 0.003$). Based on the UEQ results, the DiffusionCinema received positive evaluations ($> 0.8$) for the dimensions of Attractiveness, Perspicuity, Stimulation, and Novelty. For the RC, most metrics, e.g., Attractiveness, Perspicuity, Stimulation, Dependability, and Novelty rated as neutral (within the range of $-0.8$ to $0.8$) or Negative with its highest score being in Efficiency. 

The results indicated lower score variability in Hedonic Quality metrics between interfaces. In contrast, the Pragmatic Quality dimension did not show a pronounced difference. A potential interpretation is that users derived greater satisfaction from the contextual vusual-based interaction with generated videos compared to operating a traditional RC.

\section{Conclusion and Future Work}

We presented a new paradigm for autonomous aerial cinematography in which diffusion models translate natural language descriptions directly into executable UAV flight trajectories. By conditioning motion on textual intent and visual context, the system lets non-experts capture complex cinematic drone shots without manual piloting. Our results show that text-driven trajectories provide an intuitive interface for aerial filmmaking, with diffusion models bridging human intent and autonomous execution. A user study further validated this interaction paradigm: NASA-TLX scores revealed significantly lower overall workload for our interface ($M = 21.6$) compared to a traditional RC ($M = 58.1$), with marked reductions in mental demand ($11.5$ vs. $60.5$) and frustration ($14.0$ vs. $54.5$), underscoring the system’s usability advantages. Future work will extend the system with richer multimodal inputs, including gestures, dynamic scene understanding, and real-time constraints. We also plan to enable online trajectory refinement for adaptive cinematography and expand the dataset of cinematic prompts to improve model generalization, bringing text-to-aerial cinematography closer to practical UAV deployment.

\section*{Acknowledgements} 
Research reported in this publication was financially supported by the RSF grant No. 24-41-02039.

\bibliographystyle{ACM-Reference-Format}
\bibliography{ref}

\end{document}